\documentclass[10pt,twocolumn,letterpaper]{article}

\usepackage{cvpr}
\usepackage{times}
\usepackage{epsfig}
\usepackage{graphicx}
\usepackage{amsmath}
\usepackage{soul}
\usepackage{amssymb}
\usepackage{tabularx} 
\newcommand\Tstrut{\rule{0pt}{2.4ex}}         

\usepackage{resizegather}
\usepackage{multirow}
\usepackage{caption,subcaption}
\usepackage{graphicx,subcaption}

\usepackage[export]{adjustbox}

\DeclareMathOperator*{\argmin}{arg\,min}

\makeatletter
\def\thickhline{%
  \noalign{\ifnum0=`}\fi\hrule \@height \thickarrayrulewidth \futurelet
   \reserved@a\@xthickhline}
\def\@xthickhline{\ifx\reserved@a\thickhline
               \vskip\doublerulesep
               \vskip-\thickarrayrulewidth
             \fi
      \ifnum0=`{\fi}}
\makeatother

\newlength{\thickarrayrulewidth}
\usepackage[pagebackref=true,breaklinks=true,letterpaper=true,colorlinks,bookmarks=false]{hyperref}

\cvprfinalcopy 

\def\cvprPaperID{7473} 

\ifcvprfinal\fi
\begin{document}

\title{Training Multi-bit Quantized and Binarized Networks with\linebreak A Learnable Symmetric Quantizer}

\author{Phuoc Pham \\
 Incheon National Univeristy \\
{\tt\small phuocphn@inu.ac.kr}
\and
Jacob A. Abraham \\
University of Texas at Austin\\
{\tt\small jaa@cerc.utexas.edu}
\and
Jaeyong Chung \\
Incheon National Univeristy\\
{\tt\small jychung@inu.ac.kr}
}
\maketitle

\begin{abstract}
Quantizing weights and activations of deep neural networks is essential for deploying them in resource-constrained devices, or cloud platforms for at-scale services. While binarization is a special case of quantization, this extreme case often leads to several training difficulties, and necessitates specialized models and training methods. As a result, recent quantization methods do not provide binarization, thus losing the most resource-efficient option, and quantized and binarized networks have been distinct research areas. We examine binarization difficulties in a quantization framework and find that all we need to enable the binary training are a symmetric quantizer, good initialization, and careful hyperparameter selection. These techniques also lead to substantial improvements in multi-bit quantization. We demonstrate our unified quantization framework, denoted as UniQ, on the ImageNet dataset with various architectures such as ResNet-18,-34 and MobileNetV2. For multi-bit quantization, UniQ outperforms existing methods to achieve the state-of-the-art accuracy. In binarization, the achieved accuracy is comparable to existing state-of-the-art methods even without modifying the original architectures.
\end{abstract}

\section{Introduction}
\begin{figure}[ht]
\centering
\includegraphics[width=0.48\textwidth]{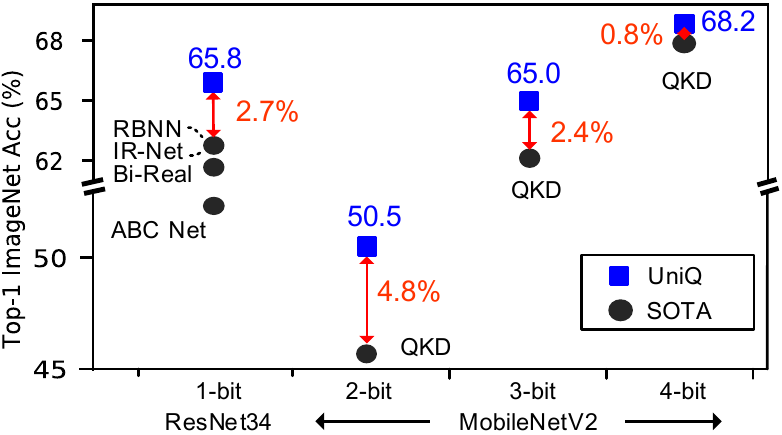}
\caption{The proposed unified quantization method outperforms the existing state-of-the-art methods both in binarization and multi-bit quantization by significant margins.}\label{main_results}
\end{figure}

Deep neural networks have achieved tremendous success in various fields including computer vision~\cite{krizhevsky2017imagenet}, natural language processing~\cite{vaswani2017attention}, and speech recognition~\cite{dahl2011context}, having demonstrated unprecedented predictive performance. However, the computational complexity and memory access count required by the existing models pose a challenge in deploying them to resource-constrained devices, and applying to latency-critical services. To address this challenge, efficient network architectures, manually designed~\cite{sandler2018mobilenetv2} or automatically searched~\cite{tan2019mnasnet}, and model compression techniques such as pruning~\cite{han2015learning} and quantization~\cite{choi2018pact,jacob2018quantization,sze2017efficient,zhang2018lq} have been studied.  In practical model deployment, quantization is a necessary step, and often the last means to control the performance and efficiency trade-off once the model architecture is fixed~\cite{jacob2018quantization}. 

Quantizing weights and activations to a lower precision not only reduces the computational complexity but also the model size, memory footprint, and memory access count but at cost of degraded performance. Recent quantization methods~\cite{lsq,kim2019qkd} can overcome accuracy degradation from their full-precision counterparts even with 4-bit weights and activations. However, most of these methods do not present 1-bit results, possibly due to their severely degraded performance or because 1-bit training does not converge, ditching the most efficient option.

While Zhou \etal~\cite{zhou2016dorefa} considered binarized neural networks together with multi-bit quantization, binarized neural networks have been a distinct research topic from the quantized models. Most studies on binarized models~\cite{courbariaux2015binaryconnect,hubara2016binarized,liu2018bi,rastegari2016xnor} focused on the 1-bit case solely. Binarized networks have gained attention because of the expensive floating-point multiplications and additions being replaced by efficient \textsc{xnor} and \textsc{popcount} operations. However, this is not specific to binarized networks, and any low-precision networks can be executed efficiently by such operations as shown in~\cite{zhou2016dorefa}. Thus, a binary kernel can execute any bit-width networks seamlessly depending on the accuracy and efficiency requirements.

In contrast, the process of building binarized networks is different from building quantized networks. Binarized network researchers often modify a base architecture to improve performance. Several studies increased the representation capacity by using more weight and activation bases~\cite{lin2017towards, zhuang2019structured}. Most studies incorporate changes to improve training efficiency such as dual skip connections~\cite{he2020proxybnn,liu2018bi,ye2020distillation}. In addition, more aggressive changes are sought for via neural architecture search ~\cite{kim2020learning,phan2020binarizing, ye2020distillation}.
In addition to model changes, binarized networks use a quantizer specialized for binary such as the sign function. 

The distinct creation process for binarized networks causes several difficulties in practice.  First, modifications of a base model often affect the number of floating-point operations and memory access count~\cite{ding2019regularizing}. Thus, the actual latency and power consumption can differ from the expected results, and a binarized model cannot be guaranteed to be better than the 2-bit counterpart. Second, training binarized networks requires additional skilled workforce because it needs special expertise, especially a deep understanding of the model itself. Finally, given a model, binarization provides only one option for accuracy and efficiency, and the limited exploration of the trade-off may lead to a suboptimal solution.

\begin{figure}[t]
\centering
\includegraphics[width=0.48\textwidth]{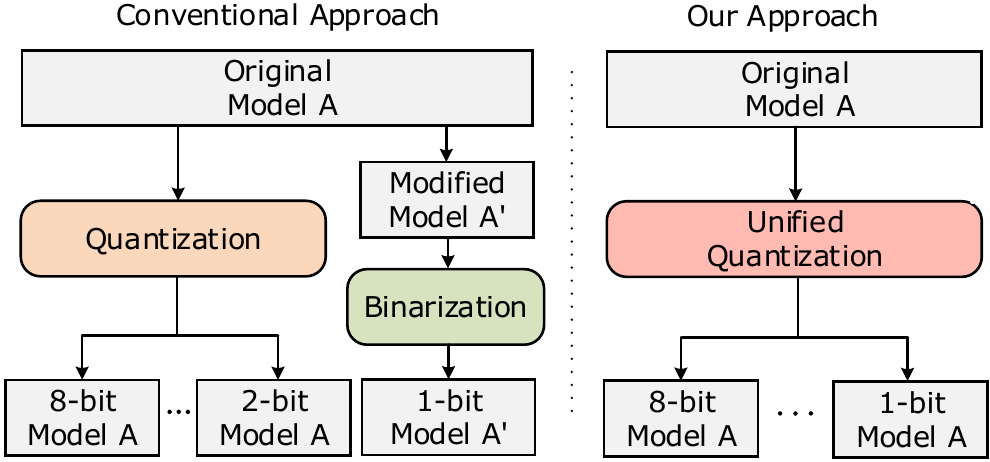}
\caption{The overview about our method. A unified framework is used for both model quantization and binarization.}\label{overview}
\end{figure}

This paper proposes a unified quantization framework, denoted as UniQ, for multi-bit quantized networks and binarized networks. UniQ achieves up to \textbf{4.8}\% and \textbf{2.7}\% accuracy improvements on the ImageNet dataset over the existing state-of-the-art quantization and binarization methods, respectively, as shown in Figure~\ref{main_results}. Figure~\ref{overview} illustrates UniQ in contrast to the conventional approach. 

This paper first considers two popular weight quantizers in quantized networks. A weight quantizer, first proposed in~\cite{zhou2016dorefa} and adopted in~\cite{choi2019accurate,jinneural} later, maps weight values into the range [0, 1] first and then performs quantization and re-maps the quantized values to the range [-1,1]. This method implies the importance of the symmetry; however, the weight mapping becomes an impediment to taking advantage of pre-trained models. The other weight quantizer, used in~\cite{lsq+, lsq}, maps inputs to a real value represented by the product of a scaling factor and a signed integer. The signed integer is assumed to be represented by two's complement, which has asymmetric ranges. This quantization method does not transform the weights of the pre-trained models. However, we hypothesize that the asymmetry has a negative impact on extremely low-precision training. Thus, we design a symmetric quantizer, where the step size can be learned via the gradient descent procedure.
 
 While the step size is learnable with the task loss, we observe that the initialization of the learnable parameter has a significant impact on the final solution. For the initialization of this parameter, prior works used a heuristic~\cite{lsq} or a numerical method~\cite{lsq+}. In contrast, we propose an analytic initialization method that is optimal in the mean squared sense. According to our ablation study, our proposed framework shows significant improvements over prior works as a combined result of the symmetric quantizer and the optimal initialization.
 
 The proposed quantizer and init method can be applied to the binary case seamlessly, but the significant improvements demonstrated in multi-bit are not shown in 1-bit. We scrutinize the training dynamics of the binary case and find that the binary case receives strong gradient signals at the beginning of training and the distribution of quantizer input changes extremely fast compared to multi-bit cases. We hypothesize that this difference is caused as  the initial point after binarization is too far from the pre-trained model solution. As a simple yet effective solution, we suggest to use warm-up strategy, which has been used widely for large-batch training. However, we empirically show that in 1-bit training, this improves accuracy substantially even when a small batch size is used.
 
 Our major contributions are summarized as follows:
 \begin{itemize}
 \item In multi-bit quantization, the proposed \textit{unified} method outperforms existing methods consistently to achieve the state-of-the-art accuracy of ResNet-18,-34 and MobileNetV2 on ImageNet. 
 \item In binarization, the proposed method achieves comparable results to the state-of-the-art methods. These results have been achieved only by enhancing the training process \textit{without modifying} the original network architectures, meaning that our method can be used in conjunction with network modification techniques. 
 \item
We propose an \textit{optimal, analytic} initialization for step sizes. 
\end{itemize}

\section{Related Work}
Modern neural networks have increased their computational complexity and memory requirements. Therefore, recent works have proposed efficient architectures~\cite{howard2019searching, howard2017mobilenets, sandler2018mobilenetv2, efficientnet} and model compression techniques such as  network binarization~\cite{courbariaux2015binaryconnect,hubara2016binarized,liu2018bi,rastegari2016xnor}, low-bit quantization~\cite{choi2018pact,jacob2018quantization,sze2017efficient,zhang2018lq}, and knowledge distillation~\cite{kim2019qkd, polino2018model} to reduce the model size and amount of computation.  Among these, low-bit quantization is one of the most popular methods and is widely used in the research literatures and real-life applications~\cite{conti2018xnor, liu2020aqd,umuroglu2017finn}. 

\textbf{Efficient Models.} Recently optimized networks such as EfficientNet~\cite{efficientnet}, MobileNet-v1~\cite{howard2017mobilenets}, -v2~\cite{sandler2018mobilenetv2}, -v3~\cite{howard2019searching} have achieved high accuracy by replacing the standard convolutional layers with depth-wise separable convolutions, thereby significantly reducing the number of parameters. Even for such efficient architectures, quantization is necessary to deploy them in specialized hardware~\cite{genc2019gemmini} and provides further reductions in its size and number of calculations.   Recent works~\cite{gong2019differentiable,jain2019trained,kim2019qkd} attempted to quantize these models, but at the expense of the significant loss in prediction accuracy. Compared to  DSQ~\cite{gong2019differentiable} and QKD~\cite{kim2019qkd}, our method yields consistently higher results for all bit-widths when tested with MobileNet-V2, which again, demonstrates its effectiveness even for highly optimized networks. 

\textbf{Model Binarization.} As a special case of quantization, model binarization has been studied extensively and has received much attention owing to its efficiency for deployment in edge devices. Using binarized weights and activations resulted in $\sim$32× memory saving over the full-precision counterpart and brought $\sim$58× computational efficiency on CPUs by taking advantage of bitwise operations~\cite{rastegari2016xnor}. Unfortunately, these networks usually lead to severe accuracy degradation. To mitigate this problem, many existing methods proposed the idea of modifying the original architecture. In~\cite{rastegari2016xnor}, the order of layers within a block was changed to improve the information flow. In~\cite{liu2018bi}, an additional skip connection was added to each block in the residual networks. In~\cite{Kim2020BinaryDuo:}, the input and output widths of each layer were adjusted. Rather than modifying the original architecture, this study focuses only on improving the quantizer itself and the training process. Our proposed method is orthogonal to the aforementioned model modification methods. 

\textbf{Multi-bit Quantization.} In contrast, recent works on quantization~\cite{choi2018pact,esser2019learned, gil2019learning,uhlich2019differentiable, wang2019haq, zhang2018lq} have achieved substantial efficiency improvements without the need to re-design or develop the whole new architecture.  Thus, it can substantially reduce the design effort. 
 PACT~\cite{choi2018pact} and LQ-Nets~\cite{zhang2018lq} first proposed the idea of learning quantizer parameters. LQ-Net parameterizes quantization levels for a non-uniform quantizer. In QIL~\cite{gil2019learning}, a non-uniform quantizer was constructed using a non-linear transformer followed by a uniform quantizer. While these non-uniform quantizers provide a higher degree of freedom, they usually require more computation and memory than uniform quantizers. PACT~\cite{choi2018pact} parameterized the clipping level in a uniform activation quantizer. LSQ~\cite{lsq} showed better accuracy by making the step size learnable. SAT~\cite{lsq} studied efficient training for quantized networks. Both PACT~\cite{choi2018pact} and SAT~\cite{choi2018pact} adopted the weight quantizer of DoReFaNet~\cite{zhou2016dorefa}, which is symmetric but  transforms the weights into a new range. This makes it difficult to take advantage of pre-trained models. In contrast, we design a uniform, symmetric quantizer that does not require the transformation and can leverage pre-trained models fully. 
Besides, none of the recent methods reports their results for binarized neural networks, due to severe performance degradation, or their quantizers are not properly designed to support binarized networks. In contrast, we propose a unified framework that can support all bit-widths including 1-bit binarization. We obtain new state-of-the-art results for multi-bit quantization and promising results for model binarization by using the new quantizer design without any modifications in the original architecture needed.

\textbf{Knowledge Distillation.} Another popular method is knowledge distillation that is widely used in many computer vision tasks. The basic idea is that the knowledge from the teacher networks is transferred to the student networks, providing an additional guidance signal to the training process of the smaller-sized student network. Applying distillation methods to low-precision networks was performed by~\cite{kim2019qkd,mishra2017apprentice,polino2018model,wu2016binarized} where a real-valued network is used as the teacher and a low-precision bit network as a student. QKD~\cite{kim2019qkd} reported competitive results in multi-bit quantization using knowledge distillation. LSQ~\cite{lsq} also showed that knowledge distillation provided additional improvements in their quantization results.  However, we outperform these methods even without resorting to the idea of transferring knowledge, by focusing more on the fundamental issues. In addition, our method is orthogonal to knowledge distillation, and can be used in conjunction to further boost the performance.

\section{Preliminaries}
In this section, we first review the weight quantizers commonly used in the literature. In~\cite{lsq+, lsq,lin2016fixed}, a weight $w \in \mathbb{R}$ is approximately represented by 
\begin{equation} 
w \approx \hat w \cdot \Delta  
\end{equation}
where $\Delta$ is a scaling factor, called the step size, and $\hat w$ is a signed integer, which is assumed to be represented by two's complement. 
This is often referred to as the fixed-point representation.  This representation has asymmetric ranges; it can represent one more negative number than positive numbers. For example, 
when $\Delta=1$, it can represent  -2, -1, 0, and 1 for 2-bit weights.     We hypothesize that this asymmetry has a negative impact on low-precision training. While it has an asymmetric range in the strict sense, it is considered  symmetric in~\cite{lsq+}. Thus, to avoid confusion, we refer to this as \textit{semi-symmetric}.
 
In~\cite{choi2019accurate, jinneural, zhou2016dorefa}, a weight $w\in \mathbb{R}$  is quantized into a $k$-bit value by \begin{equation}
2 \text { Quantize }_{\mathrm{k}}\left(\frac{\tanh \left(w\right)}{2 \max \left(\left|\tanh \left(w\right)\right|\right)}+\frac{1}{2}\right)-1
\end{equation}
where $\text{ Quantize }_{\mathrm{k}}(x) = \text{round}(x (2^k-1) )/(2^k-1)$. In this quantizer, the weights are first mapped into the range [0,1] and quantized into a $k$-bit value in the range [-1,1]. This quantizer has the symmetric property, but is problematic for two reasons. First, it transforms the weights into a new range and loses the knowledge of pretrained models. Second, the transformed range may cause the vanishing or exploding gradient problem because the variance of weights becomes substantially different from that suggested in Xavier~\cite{glorot2010understanding} or Kaiming initialization~\cite{he2015delving}. Thus, SAT~\cite{jinneural} proposed to scale the transformed weights again using the constant in Xavier initialization. However, the first problem remains. We address these two problems by using a new symmetric quantizer and optimal initialization that minimizes the mean square quantization error. 
\section{Unified Quantization}

\subsection{Learnable Symmetric Quantizer}
A quantizer $Q:\mathbb{R}\rightarrow \mathbb{R}$ is a piecewise constant function and each interval is mapped to a corresponding output. The end points of the intervals are referred to as \textit{decision levels} and the output is called the \textit{reconstruction level}. A \textit{uniform} quantizer has evenly spaced decision levels and reconstruction levels.  The length of the intervals is called the \textit{step size}, denoted as $\Delta$. The total number of reconstruction levels is denoted by $N$. We denote the clip function to be used by quantizers by $\text{clip}_N(x)=\min(\max(x, 0), N-1)$. The round function is denoted by $\lfloor \cdot \rceil$. In practice, $N$ is usually even, and thus a symmetric quantizer in the strict sense do not include the value of zero as a reconstruction level.  For example, when $N=4$ and $\Delta=1$, the symmetric quantizer has -1.5, -0.5, 0.5, and 1.5 as the reconstruction levels. We quantize weights by the uniform symmetric quantizer    
\begin{equation}\label{wq}
  Q_w(x) =  \lfloor \text{clip}_N\left(\left(x+\alpha\right)/\Delta \right )\rceil\Delta - \alpha
\end{equation}
where $\alpha = \Delta\cdot (N-1)/2$. Eq. (\ref{wq}) can be rewritten as
\begin{equation}
Q_w(x) =   \hat q_w (\Delta/2)
\end{equation}
where $\hat q_w=2\lfloor \text{clip}_N\left( (x+\alpha)/\Delta\right )\rceil-N+1 $; $\hat q_w$ can be encoded into $\lceil \log_2{N} \rceil$ bits using $\pm1$ encoding.

We consider the ReLU non-linearity, wihch is widely used in the deep learning literature, as the activation function. Because almost half of the ReLU responses are zero, we fix the zero value as a  reconstruction level instead of parameterizing an offset.  Thus, for activations, we use  
\begin{equation}\label{aq}
Q_a(x) = \lfloor \text{clip}_N( x/\Delta)\rceil\Delta.
\end{equation}

Designing a uniform quantizer usually boils down to deciding one parameter, the step size $\Delta$. Instead of designing the quantizers manually, we make $\Delta$ a learnable parameter as in recent prior works~\cite{lsq,lsq+},  and optimize it with the task loss via the gradient descent procedure. The round function has a zero derivative almost everywhere, and the exact derivative is not useful in learning. Thus, we adopt the straight-through estimator (STE), which assumes a unit derivative for the entire input range of the round function. Then, we have
\begin{equation}
\frac{\partial Q_w}{\partial \Delta}=\left\{\begin{array}{ll}
-\frac{x}{\Delta} + \lfloor \frac{x}{\Delta}-\frac{1}{2} \rceil + \frac{1}{2} & \text { if }|x| <\alpha \\
\text{sign}(x)\alpha & \text { otherwise. }\\ 
\end{array}\right.
\end{equation}
We can also find $\frac{\partial Q_a}{\partial \Delta}$ similarly.
While this allows us to learn the step size, the initialization of this parameter is necessary and in our experience, careful initialization improves accuracy substantially.
\subsection{Optimal MSE Initialization}

The quantized networks are usually initialized with a pre-trained model, and the learnable step sizes are also initialized depending on the statistics of the pre-trained model.  Let $X$ be the random variable for a quantizer input and its pdf is denoted by $p(x)$. The optimal step size for $Q_w$ is defined in the mean squared error (MSE) sense by
\begin{equation}
\Delta^*_w = \argmin_{\Delta} D_w(\Delta)
\end{equation}
 where $D_w(\Delta)=\mathop{\mathbb{E}}[(x-Q_w(x;\Delta))^2]$. 
 Using Leibniz integral rule, we take the derivative of $D_w$ and obtain 
\begin{equation}\label{dw}
\begin{aligned}
\frac{d D_w}{d \Delta}=-\sum_{i=1}^{\frac{N}{2}-1}(2 i-1) \int_{(i-1) \Delta}^{i \Delta} 2\left(x-\left[\frac{2 i-1}{2}\right] \Delta\right) p(x) d x \\
-(N-1) \int_{(\frac{N}{2}-1) \Delta}^{\infty} 2\left(x-\left[\frac{N-1}{2}\right] \Delta\right) p(x) d x.
\end{aligned}
\end{equation}


 By setting (\ref{dw}) to zero, we can find the optimal step size. 
In general, this equation does not have a closed-form solution and a numerical method is required. However, for common probability distributions such as Gaussian and Laplace, the step size for each $N$ of interest can be pre-computed assuming a unit variance, and can be scaled by the standard deviation of the quantizer input.  In our implementation, a Gaussian distribution is assumed for weights.

For the activation quantizer $Q_a$, we also define the optimal step size $\Delta_a^*$ and the mean squared error $D_a$, and derive $\frac{dD_a}{d_\Delta}$ similarly. However, in the case of the activation quantizer, a Gaussian distribution is assumed for \textit{pre-activations} (activations prior to the non-linearity). The activations after the ReLU non-linearity follow a rectified Gaussian, a modification of Gaussian where the negative elements are reset to zero.  While a rectified Gaussian is a mixture of a discrete distribution for zero and a continuous distribution for the positive elements, we pre-compute the step size using the continuous part only because $Q_a$ is designed to include zero as a reconstruction level by construction. When we pre-compute the step sizes, the standard Gaussian is assumed for pre-activations. We denote the pre-computed step sizes for activations and weights by $\Delta^u_a$ and $\Delta^u_w$, respectively. 
\begin{table}[]

	\caption{The optimal unit step size and optimal SQNR for weights and activations.}\label{unitstep}
	\centering

	\begin{tabular}{c|c|c|c|c}
		\thickhline
	\multirow{2}{*}{$N$}	&   \multicolumn{2}{c|}{Weight} & \multicolumn{2}{c}{Activation}\\
	\cline{2-5}
	     & $\Delta^u_w$ & SQNR(~\textit{dB}) & $\Delta^u_a$ & SQNR(~\textit{dB}) \\
\hline
		2 & 1.596	& 4.4  &  1.224 & 5.5	\\
		4& 0.996 	& 9.3  &  0.651 & 11.6\\
		8& 0.586 	& 14.3 &   0.353 & 17.2	 \\
		16& 0.335 	& 19.4 &	0.193 &	22.7 \\
	\thickhline

	\end{tabular}
\end{table}
The constants for each $N$ are summarized in Table~\ref{unitstep}, which also shows the optimal signal-to-quantization-noise (SQNR) ratio. We use it later to analyze the training dynamics of quantized models. Even if the step size is set optimally, the MSE is proportional to the signal energy (the variance of the quantizer input) and it is not useful to see the optimality of the step size during training where the signal energy varies substantially over time.  Using the pre-computed step sizes, we finally obtain 
\begin{equation}\label{w_init}
\Delta^*_w = \Delta^{u}_w \text{Std}(X), 
\end{equation}
and
\begin{equation}\label{a_init}
\Delta^*_a = \Delta^{u}_a \sqrt{2 \mathop{\mathbb{E}}[X^2]}.
\end{equation}
Note that $\sqrt{2\mathop{\mathbb{E}}[X^2]}$ is the standard deviation of the pre-activations because they were assumed to have a symmetric distribution around zero. The statistics of the quantizer input are estimated by the sample statistics. For activations, we use a given number of batches to estimate the standard deviation. For a simple implementation, we calculate the sample standard deviation of each batch and use its maximum values over the batches. 
In order to accurately estimate the input statistics of an activation quantizer using multiple batches, we need to forward-propagate multiple batches for each layer in a layer-wise manner. In our experience, this adds unnecessary complexity to the implementation and the simple method provides similar performance.

\subsection{Training Instability in 1-bit}
While our symmetric weight quantizer and MSE init support all bits seamlessly to 1-bit, in our experience, the binary training is not effective in the same setup as that for other bits. We investigate the training dynamics,can which leads to the following observations. First, 1-bit SGD training receives strong gradient signals initially because the initial point after binarization is far from the solution of the pre-trained model, which we use for init, in contrast to 2-bit or higher training. Second, the step sizes are not adapted to maximize the signal-to-quantization-noise ratio (SQNR) or maintain the initial SQNR during the initial fast learning. While the objective of the learnable quantizer is not to maximize the SQNR, we observe that the step size usually changes along with the standard deviation of the quantizer input in 2-bit or higher training, maintaining a reasonable SQNR. Thus, a SQNR significantly lower than the optimal level is not considered as desirable.  We hypothesize that abrupt changes in the quantizer input distribution get the step size stuck in a local minimum. We show empirical evidence that warm-up training mitigates this issue. In addition, we empirically find that Adam is more robust to this problem than SGD. This appears to be owing to the gradient normalization in Adam. 


\section{Experimental Results}
To demonstrate the effectiveness of our proposed method, we evaluate it on the CIFAR-100~\cite{krizhevsky2009learning} and the ImageNet datasets~\cite{imagenetdataset}. The CIFAR-100 dataset consists of 60,000 32x32 color images from 100 classes with a total of 50,000 training and 10,000 test images. The ImageNet dataset consists of more than 1.2M training images from 1000 classes and 50K validation images. We use various popular network architectures such as ResNet-18, -32, -34~\cite{he2016deep} and MobieNet-V2~\cite{sandler2018mobilenetv2} for evaluation. The experiment results are compared with various recent works on multi-bit quantization and neural network binarization.


\begin{table*}[t]
\begin{center}
\begin{tabular}{l|cccc|cccc|cccc}
\thickhline
\multicolumn{1}{c}{\multirow{3}{*}{Method}} & \multicolumn{4}{c|}{ResNet-18 (FP: 71.57)}                     & \multicolumn{4}{c|}{ResNet-34 (FP: 75.11)}                     & \multicolumn{4}{c}{MobileNet-V2 (FP: 71.53)}                   \\ \cline{2-13} 
\multicolumn{1}{c}{}                                 & \multicolumn{12}{c}{Bit-width (W/A)}                                                                                                                                                          \\ \cline{2-13} 
\multicolumn{1}{c}{}                                 & 4/4           & 3/3           & 2/2           & 1/1           & 4/4           & 3/3           & 2/2           & 1/1           & 4/4           & 3/3           & 2/2           & 1/1           \\  \hline
PACT~\cite{choi2018pact}                                                 & 69.2          & 68.1          & 64.4          & -             & -             & -             & -             & -             & 61.4          & -             & -             & -             \\
DoReFa-Net~\cite{zhou2016dorefa}                                           & 68.1          & 67.5          & 62.6          & -             & -             & -             & -             & -             & -             & -             & -             & -             \\
DSQ~\cite{gong2019differentiable}                                                  & 69.6          & 68.7          & 65.2          & -             & 72.8          & 72.5          & 70.0          & -             & -             & -             & -             & -             \\
QIL~\cite{gil2019learning}                                                  & 70.1          & 69.2          & 65.7          & -             & 73.7          & 73.1          & 70.6          & -             & 64.8          & -             & -             & -             \\
LSQ~\cite{lsq}                                                  & 71.1          & 70.2          & 67.6          & -             & 74.1          & 73.4          & 71.6          & -             & -             & -             & -             & -             \\
LSQ+~\cite{lsq+}                                                  & 70.8          & 69.3          & 66.8          & -             & -             & -             & -             &               & -             & -             & -             & -             \\
SAT~\cite{jinneural}                                                  & 70.3          & 69.3          & 65.5          & -             & -             & -             & -             & -             & -             & -             & -             & -             \\
QKD~\cite{kim2019qkd}                                                  & 71.4          & 70.2          & 67.4          & -             & 74.6          & 73.9          & 71.6          & -             & 67.4          & 62.6          & 45.7          & -             \\ \hline
\textbf{UniQ (Ours)}                                 & \textbf{71.5} & \textbf{70.5} & \textbf{67.8} & \textbf{60.5} & \textbf{75.0} & \textbf{74.2} & \textbf{72.1} & \textbf{65.8} & \textbf{68.2} & \textbf{65.0} & \textbf{50.5} & \textbf{23.2} \Tstrut \\ \thickhline 
\end{tabular}

\end{center}
\caption{Top-1 accuracy (\%) on ImageNet dataset. Comparison with the existing state-of-the-art DSQ, QIL, LSQ, LSQ+, QKD and SAT. The result of PACT for MobileNet-V2 is cited from HAQ paper~\cite{wang2019haq}. The dash symbol "-" indicates no data available and "FP" represents the full precision network accuracy in our implementation.}
\label{tab:multibit_imagenet_result}
\end{table*}


\subsection{ImageNet Results}
\label{imagenet_result}

\textbf{Implementation details}. In the following experiments, we quantize all convolutional and fully connected layers to ultra-low precision except the first and last layers, which are represented by 8-bit precision as was done in~\cite{lsq}.  In case of binarized networks, we leave the first, last, and down-sampling layers the full-precision as were done in prior works~\cite{Kim2020BinaryDuo:, liu2018bi}. We quantize weights and activations to the same bit-width for all experiments. For multi-bit and binarized networks, we use SGD and Adam, respectively. For SGD, we use 0.01 as the initial learning rate and decay it using the cosine learning rate schedule without restarts~\cite{cosinewarmup}. For Adam, the learning rate is fixed to 0.001 for 5 epochs as the warm-up and then increase to 0.004 and then follow the cosine schedule. For all experiments, we use layer-wise and kernel-wise quantizations for activations and weights, respectively, with an exception of MobileNetV2, in which layer-wise quantization is used for both weights and activations considering the relative parameter overhead. In addition, weight decay is not used for step size parameters. Our implementation is based on PyTorch.

We use original ResNet-18, ResNet-34, and MobileNetV2 architectures, without any changes in their structure. For ResNets, we use the pre-activation version. We follow the commonly used data augmentation strategy as in~\cite{zhang2018lq,lsq}, where the training images are randomly cropped and resized to 224 × 224, and horizontally flipped half the time. For testing, the single-center crop of size 224×224 is applied. The transformed images are finally normalized by the mean and standard deviation. All networks are trained for 90 epochs with a batch size of 256 (2 GPUs), a momentum of 0.9, and a weight decay of $10^{-4}$, $0.5\times 10^{-4}$, $0.25\times 10^{-4}$, 0 for 4-bit, 3-bit, 2-bit and 1-bit quantized models, respectively. We use the pre-trained models available at PytorchCV \footnote {\url{https://pypi.org/project/pytorchcv}} for weight initialization. For multi-bit and binarized networks, we use the floating-point and 2-bit models for initialization, respectively. For step size initialization, the first 1000 training batches are used to estimate the statistics of activations. 





\begin{table}[]
\centering
\begin{tabular}{llcc}
\thickhline
Network                                                                  & Method        & Acc(\%) & Original\\ \hline
\multirow{11}{*}{\begin{tabular}[c]{@{}l@{}}ResNet-18\\ (FP: 71.57)\end{tabular}} & ABC-Net~\cite{lin2017towards}                      & 42.7             &            \\
                                                                                  & XNOR-Net~\cite{rastegari2016xnor}                    & 51.2             &             \\
                                                                                  & BNN+~\cite{darabi2018bnn+}                          & 53.0             &   $\checkmark$           \\
                                                                                  & DoReFa-Net~\cite{zhou2016dorefa}$\dag$                        & 53.4             &  $\checkmark$            \\
                                                                                  & Bi-Real~\cite{liu2018bi}                        & 56.4             &             \\
                                                                                  & XNOR-Net++~\cite{bulat2019xnor}                   & 57.1             &          \\
                                                                                  & IR-Net~\cite{qin2020forward}                          & 58.1             &      $\checkmark$        \\
                                                                                  & ProxyBNN~\cite{he2020proxybnn}                       & 58.7             &       $\checkmark$       \\
                                                                                  & RBNN~\cite{lin2020rotated}                          & 59.9             &           $\checkmark$   \\
                                                                                  & BinaryDuo~\cite{Kim2020BinaryDuo:}                       & 60.4             &            \\ \cline{2-4} 
                                                                                  & \textbf{UniQ (Ours)}            & \textbf{60.5}             &        $\checkmark$      \\ \hline
\multirow{5}{*}{\begin{tabular}[c]{@{}l@{}}ResNet-34\\ (FP: 75.11)\end{tabular}}  & ABC-Net                       & 52.4             &           \\
                                                                                  & Bi-Real                        & 62.2             &             \\
                                                                                  & IR-Net                         & 62.9             &     $\checkmark$         \\
                                                                                  & RBNN                           & 63.1             &      $\checkmark$        \\ \cline{2-4} 
                                                                                  & \textbf{UniQ (Ours)}            & \textbf{65.8}                &    $\checkmark$          \\ \thickhline
\end{tabular}
\caption{Top-1 accuracy comparison to the existing state-of-the-art binarization methods on ImageNet. "\textit{Original}" is used to denote methods not requiring architecture modification. $\dag$DoReFa-Net uses 2-bit for activations.}\label{tab:bnn_results}
\end{table}

\textbf{Comparison with prior works on multi-bit quantization.}  We compare our method to existing methods in Table~\ref{tab:multibit_imagenet_result}. For the existing methods, the results are directly cited from the original papers unless mentioned otherwise. By looking at the reported table, we can observe that our method outperforms all the previous quantization methods in top-1 accuracy. Specifically, we can achieve significant performance gain over the recent state-of-the-art methods LSQ, QKD, and SAT on all comparing architectures. The improvements range from 0.1\% to 4.8\% compared to the second-best method (QKD). Note that QKD and SAT need a total of 120 and 150 training epochs, respectively, while our method only requires 90 epochs to obtain better accuracy. In addition, it is worth to mention that QKD uses knowledge distillation but UniQ does not. Knowledge distillation is known to provide additional improvements on quantization results as shown in LSQ~\cite{lsq}.  MobileNetV2 has an architecture already optimized for efficiency such as depth-wise convolutions and the accuracy is known to be sensitive to quantization~\cite{lsq, kim2019qkd}. For MobileNetV2, UniQ outperforms the existing state-of-the-art method, QKD, by significant margins of 2.4\% and  4.8\% for 2-bit and 3-bit quantized models, respectively.
A substantial increase in prediction accuracy can be seen in other 2-bit models. With ResNet-18, UniQ achieves 67.8\% top-1 accuracy, with 0.2\% and 0.4\% improvements over LSQ and QKD, respectively.  With ResNet-34, it achieves a top-1 accuracy of 72.1\%, which is a 0.4\% improvement over QKD and LSQ. 

\textbf{Comparison with prior works on binarized neural networks.} We further compare our method with the state-of-the-art binarization methods in Table~\ref{tab:bnn_results}. As shown in the last column, many existing methods modify a base architecture and it is difficult to be compared. While the model size and the number of operations are the same, the memory access count can be different. For example, the dual skip connection employed in Bi-Real does not affect the model size and the number of operations for convolutions but the memory access count. Nonetheless, to our best knowledge, UniQ outperforms the previous state-of-the-art accuracy for binarized ResNet-34 by a significant margin of 2.69\%. For ResNet-18, UniQ even achieves a comparable accuracy to BinaryDuo, which requires to increase the width of the skip connections.   

\subsection{CIFAR-100 Results}

\textbf{Implementation details.} We use the pre-activation variant of ResNet-32 for all the experiments on CIFAR-100. We train for 350 epochs with a mini-batch size of 128. All quantized models are initialized from the pre-trained full-precision counterparts, which we train from scratch. For simplicity, we use the same weight decay of $5\times 10^{-4}$ across all CIFAR-100 experiments. The standard data augmentation includes random cropping and horizontal flipping is applied for each training image. The first 100 training batches are used for step size initialization. We use layer-wise quantization for both weights and activations.  For other settings, we follow the same settings as described in Section \ref{imagenet_result}

\textbf{Comparison with LSQ.} 
For a fair comparison, we compare LSQ to our method in our setting. We implement LSQ carefully and cross-check the correctness with~\cite{lsq,lsq+}. Our final results are summarized in Table ~\ref{tab:lsq_uniq_cifar100_results}. When using ResNet-32, we can observe that, for 4-bit quantized models, LSQ can match the accuracy of the full-precision baseline, which is in line with the results reported in the original paper~\cite{lsq}. It is worth noting that when the bit-width is reduced to 3, our method can still archive the same accuracy of 71.4\% compared to the 4-bit LSQ quantized model. For  2-bit, the accuracy drops by only 2.1\% when using our method compared to 2.9\% for LSQ. For the most aggressive 1-bit quantization, we can achieve 62.4\%, while no data is reported for LSQ  as its quantizer is not suitable for binarization.

\begin{figure}[h]
\centering
\includegraphics[width=0.48\textwidth]{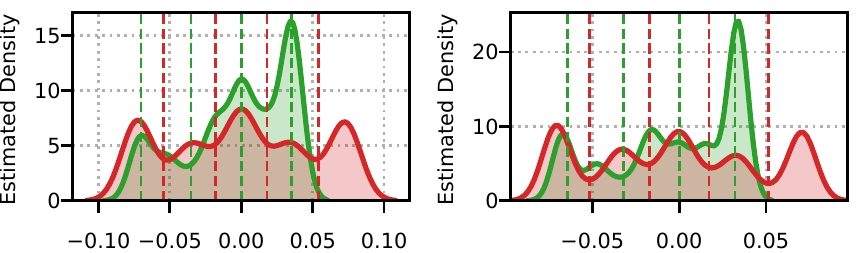}
\caption{Weight distribution from two different layers of trained, 2-bit quantized ResNet-32 with LSQ (green) and UniQ (red) quantization scheme. Two networks are trained with the same hyperparameters on CIFAR-100 for 350 epochs. The dashed lines are the reconstruction levels of each method.}
\label{fig:weight_distribution}
\end{figure}

\begin{figure*}[ht]
\centering
\begin{subfigure}[b]{.24\textwidth}
\includegraphics{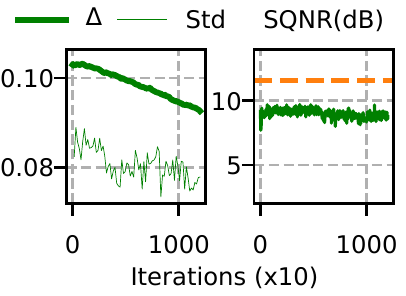}
\caption{2-bit SGD.}
\label{fig:training_dynamics_a}
\end{subfigure}
\begin{subfigure}[b]{.24\textwidth}
\includegraphics{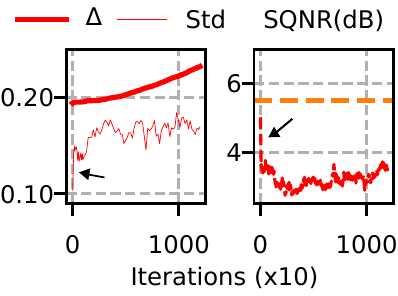}
\caption{1-bit SGD w/o warm-up.}
\label{fig:training_dynamics_b}
\end{subfigure}
\begin{subfigure}[b]{.24\textwidth}
\includegraphics{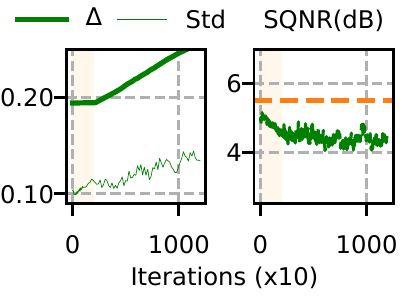}
\caption{1-bit SGD w/ warm-up.}
\label{fig:training_dynamics_c}
\end{subfigure}
\begin{subfigure}[b]{.24\textwidth}
\includegraphics{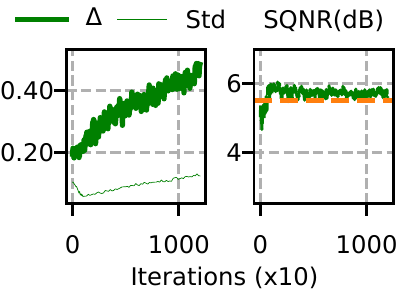}
\caption{1-bit Adam.}
\label{fig:training_dynamics_d}
\end{subfigure}

\caption{Training dynamics of an activation quantization layer in ResNet32. The abrupt change of the standard deviation (std) of the quantizer input may get the step size ($\Delta$) stuck in a poor local minimum. The highlighted area indicates the warm-up period and the dashed orange line indicates the optimal SQNR value.}
\end{figure*}

\textbf{Imbalanced weight distribution. } For UniQ and LSQ, we show the distributions of the trained weights in  Figure~\ref{fig:weight_distribution}, for two different layers. We can see that the trained weights of LSQ have the form of negatively-skewed distribution, with a long tail on the negative side, and many weights values around the maximum reconstruction level. In contrast, UniQ has a  symmetric distribution around the zero value, and the weights are relatively evenly distributed. The higher entropy of the balanced distribution may suggest that UniQ allows the network to retain more knowledge on weights than LSQ.

\textbf{Step size initialization.} To show how important the initialization of the step sizes is, we also compare the results when the step size of the proposed quantizer is initialized with 0.1, 0.2, LSQ's heuristic, and the proposed optimal method. In LSQ, the step size is initialized to \( \frac{2 \mathop{\mathbb{E}}[X]}{\sqrt{Q_P}} \) where \(X\) is the quantizer input and \(Q_P=2^{\text{bit}}-1\)   (\(Q_P=2^{\text{bit}-1}-1\)) for activations (weights). The results are summarized in Table~\ref{tab:stepsize_init_results}. The accuracy values vary substantially depending on initialization.  It is interesting to see that, our proposed quantizer with LSQ init performs poorly for all bit-widths. In contrast, by replacing LSQ init with our init, a significant performance boost can be seen for all bit-widths. The 1.0\% performance boost over LSQ init is seen in 2-bit.  These results suggest that the step size parameters need to be initialized properly, otherwise it will lead to performance degradation. 

\begin{table}[]
\begin{center}
\begin{tabular}{cccccc}
\thickhline
\Tstrut
\multirow{2}{*}{Methods} & \multicolumn{4}{c}{Bit-width (W/A)}                   \\ \cline{2-6}\Tstrut
  &32/32  & 4/4                & 3/3                & 2/2      & 1/1          \\ \hline\Tstrut
LSQ   &                                                                                 \multirow{2}{*}{71.4} & 71.4        & 70.9                   & 68.5       & -   \\ 
\textbf{UniQ (Ours)}  &                                                                                   & \textbf{71.6} & \textbf{71.4} & \textbf{69.3} &  \textbf{62.4} \\ \thickhline
\end{tabular}
\end{center}
\caption{Comparison of LSQ and our method (UniQ) with ResNet32 and CIFAR-100 (FP Acc: 71.4\%)}
\label{tab:lsq_uniq_cifar100_results}
\end{table}

\begin{table}[]
\begin{center}
\begin{tabular}{ccccc}
\thickhline
\Tstrut
Bit-width  & \multicolumn{4}{c}{Step Size  Initialization}                   \\ \cline{2-5}\Tstrut
(W/A)   &0.1  & 0.2    & LSQ Init   & Our Init              \\ \hline\Tstrut
2/2   &                                                                                 \multirow{2}{*}{ }67.1 &   68.6       &            68.3        &  \textbf{69.3}       \\ 
3/3  &70.7 & 70.9 &  71.0& \textbf{71.4}  \\ \thickhline
\end{tabular}
\end{center}
\caption{Comparison of different methods for step size initialization with ResNet32 and CIFAR-100 (FP Acc: 71.4\%)}
\label{tab:stepsize_init_results}
\end{table}



\begin{table}[]
\begin{center}
\begin{tabular}{ccccccc}
\thickhline
&\multicolumn{3}{c}{No warm-up} &\multicolumn{2}{c}{Constant warm-up} \\ \cline{2-6}
&\multicolumn{3}{c}{Learning rate}       &           \multicolumn{2}{c}{Warm-up epochs}            \\ \hline
\multirow{2}{*}{SGD} &0.01        & 0.005       & 0.001               & 5                          & 10               \\ \cline{2-6}
&57.0       & 59.1       & 59.5                & \textbf{61.3}             & 61.1            \\ \hline
\multirow{2}{*}{Adam} &0.004        & 0.001       & 0.0005               & 5                          & 10               \\ \cline{2-6}
&60.9       & 59.9       & 61.0                & \textbf{62.4}             & 62.1            \\ \thickhline
\end{tabular}
\end{center}
\caption{Results of ResNet-32 on CIFAR-100 with different learning rate schedules. For the warm-up period, the initial learning rate is set to 0.001, and then increases to 0.01 (0.004) for SGD (Adam).}
\label{tab:lr_result}
\end{table}
\raggedbottom

\textbf{Training dynamics.}  While our quantizer and init method are general to all bit-widths, the binarized networks trained by UniQ do not provide a satisfactory performance at the same hyperparameter setting for 2-bit or higher. We thus investigate the activation quantizer of a convolutional layer. Specifically, we observe the evolution of the step size, the standard deviation (SD) of the quantizer input, and SQNR during 30 epochs of training. Figure~(\ref{fig:training_dynamics_a}) and (\ref{fig:training_dynamics_b}) show 2-bit and 1-bit training, respectively, at the same setting, where we use SGD and the learning rate of 0.01. In 2-bit, the step size shrinks as the SD decreases, which maintains the initial SQNR not very far from the optimal level, shown in the dotted orange line. However, in 1-bit, the SD jumps sharply at the beginning, whereas the step size does not change accordingly, dropping the SQNR substantially, suggesting that the step size might be trapped at a local minimum.    Figure~(\ref{fig:training_dynamics_c}) shows how the warm-up can help mitigate this problem. We use 5 epochs of warm-up with a learning rate of 0.001 and observe that the SD changes smoothly and the step size seems to be adapted to that, maintaining a better SQNR.   Figure~(\ref{fig:training_dynamics_d}) shows an evidence that Adam is more robust to this issue. While we use the learning rate of 0.004, which is a high rate for Adam, we do not observe the issue, and the SQNR approximately stays at the optimal level in this case.

\textbf{Warm-up for 1-bit Training.} 
We further perform experiments to find the best optimizer and warm-up strategy for binary training. 
The empirical results from Table~\ref{tab:lr_result} are in line with our analysis on the training dynamics. The results indicate that initially training a binarized model with a small learning rate for some epochs can improve the prediction accuracy substantially. In contrast, the models trained with a higher rate at the beginning can be easily trapped in a saddle point or bad local minimum; without increasing the learning rate at some point, the training may converge too slow. The results also suggest that Adam may perform better than SGD, but SGD also provides a decent performance in contrast to the common belief that SGD gives severely degraded results than Adam in binary training~\cite{alizadeh2018empirical}. It is also shown that warm-up training helps Adam as well as SGD.   Moreover, we observe that increasing the number of warm-up epochs does not improve the accuracy significantly. Thus, we choose the 5 epoch warm-up period. 
\section{Conclusion}
In this paper, we have proposed a quantization method generalized for both multi-bit quantized and binarized models. We have designed a symmetric quantizer with a trainable step size and proposed an analytic, optimal initialization of the step size. In addition, we have investigated the difficulties in the 1-bit training and suggested practical methods to overcome them. For multi-bit quantization, the proposed method have achieved new record accuracies of ResNet-18,-34, MobileNetV2 on ImageNet. For binarization, without modifying original network architectures, we have achieved better or comparable accuracy to that of recent binarized networks by focusing on the fundamental training problems.
{\small
\bibliographystyle{ieee}
\bibliography{egbib}
}

\end{document}